\documentclass[runningheads]{llncs}
\usepackage{graphicx}
\usepackage{booktabs}
\usepackage[colorinlistoftodos]{todonotes}
\usepackage[inline]{enumitem}

\usepackage{chngcntr}
\counterwithout{figure}{chapter}
\counterwithout{table}{chapter}

\begin{document}
%
\title{Creating and Querying Personalized Versions of Wikidata on a Laptop}
%
%
\author{Hans Chalupsky\inst{1} \and Pedro Szekely\inst{1} \and Filip Ilievski\inst{1}  \and \\ Daniel Garijo\inst{2}   \and   Kartik Shenoy\inst{1} }
\authorrunning{Chalupsky et. al.}
%
\institute{Information Sciences Institute, University of Southern California \email{\{hans,pszekely,ilievski,kshenoy\}@isi.edu} 
\and Ontology Engineering Group, Universidad Politécnica de Madrid \\ \email{daniel.garijo@upm.es}}
\maketitle              
\setcounter{footnote}{0}
\setcounter{table}{0}

\begin{abstract}\let\thefootnote\relax\footnotetext{Copyright © 2021 for this paper by its authors. Use permitted under Creative Commons License Attribution 4.0 International (CC BY 4.0).}

Application developers today have three choices for exploiting the knowledge present in Wikidata: they can download the Wikidata dumps in JSON or RDF format, they can use the Wikidata API to get data about individual entities, or they can use the Wikidata SPARQL endpoint. None of these methods can support complex, yet common, query use cases, such as retrieval of large amounts of data or aggregations over large fractions of Wikidata.
This paper introduces \texttt{KGTK Kypher}, a query language and processor that allows users to create personalized variants of Wikidata on a laptop. We present several use cases that illustrate the types of analyses that Kypher enables users to run on the full Wikidata KG on a laptop, combining data from external resources such as DBpedia. The Kypher queries for these use cases run much faster on a laptop than the equivalent SPARQL queries on a Wikidata clone running on a powerful server with 24h time-out limits.

\textbf{Type:} Research paper
 
\end{abstract}

\section{Introduction}


Modern Knowledge Graphs (KGs) are increasingly focused on improving their coverage of instances and statements and enhancing their expressivity in order to support application needs such as question answering and entity linking. As a result, Wikidata~\cite{vrandevcic2014wikidata}, a popular and representative Knowledge Graph, contains nearly 95 million entities described with over 1.3 billion statements.\footnote{\url{https://grafana.wikimedia.org/d/000000175/wikidata-datamodel-statements}}
Wikidata is also highly expressive, using a reification model where each statement includes qualifiers (e.g., to indicate temporal validity) and references (which provide the source(s) from which the statement comes from). 

Application developers today have three choices for exploiting the knowledge present in Wikidata. They can download the Wikidata dumps in JSON or RDF \cite{rdf} format, they can use the Wikidata API to get data about individual entities, or they can use the Wikidata SPARQL \cite{sparql}  endpoint for more elaborate and complex queries.\footnote{\url{https://query.wikidata.org/}} 
The public Wikidata SPARQL endpoint restricts queries to 5 minutes, returning an error when a query exceeds (or plans to exceed) that time. To mitigate the time-limit restriction, developers can load the massive Wikidata RDF dump on their own servers, a relatively complex process that requires a large server and several days. 

This paper introduces \texttt{Kypher}, the query language and processor of the KGTK Knowledge Graph Toolkit~\cite{ilievski2020kgtk}, which allows creating personalized variants of Wikidata on a laptop, and enables running analytic queries faster than a Wikidata SPARQL endpoint. Because Kypher uses the KGTK representation, it is not restricted to Wikidata, and can be used to query RDF KGs such as DBpedia~\cite{auer2007dbpedia}.
The key advantages of Kypher over existing tooling are: 
\begin{enumerate}
    \item Ability to extract large amounts of data from Wikidata.
    \item Ability to execute queries that retrieve large portions of the full Wikidata.
    \item Ability to build personalized versions of Wikidata, and extending it with other datasets for specific use cases.
    \item Easy installation as there are no databases to set up or administer.
    \item Minimal hardware requirements, as Kypher can be used to query Wikidata and DBpedia on a laptop.
\end{enumerate}


The rest of the paper is structured as follows. 
Section~\ref{sec:background} describes the KGTK toolkit, and section~\ref{sec:kypher} presents an  introduction to the Kypher query language.
Section~\ref{sec:usecases} introduces five representative use cases to illustrate the benefits of Kypher, and section~\ref{sec:experiments} reports the times needed to address the use cases using Kypher queries on a laptop; SPARQL queries on a clone of the Wikidata endpoint; and SPARQL queries on on the public Wikidata endpoint. Section~\ref{sec:discussion} presents conclusions, discussion of the results and directions for future work.


\section{Background}
\label{sec:background}


The Knowledge Graph Toolkit (KGTK)~\cite{ilievski2020kgtk} is a comprehensive framework for the creation and exploitation of large hyper-relational KGs, designed for ease of use, scalability, and speed. KGTK represents KGs in tab-separated (TSV) files with four columns: edge-identifier, head, edge-label, and tail. All KGTK commands consume and produce KGs represented in this format, so they can be composed into pipelines to perform complex transformations on KGs. KGTK provides a suite of import commands to import Wikidata, RDF and popular graph representations into the KGTK format. A rich collection of transformation commands make it easy to clean, union, filter, and sort KGs; graph combination commands support efficient intersection, subtraction, and joining of large KGs; 
graph analytics commands support scalable computation of centrality metrics such as PageRank, degrees, connected components and shortest paths; advanced commands support lexicalization of graph nodes, and computation of multiple variants of text and graph embeddings over the whole graph. In addition, a suite of export commands supports the transformation of KGTK KGs into commonly used formats, including the Wikidata JSON format, RDF triples, JSON documents for ElasticSearch indexing and  graph-tool.\footnote{\url{https://graph-tool.skewed.de}} Finally, KGTK allows browsing KGs  in a UI using a variant of SQID;\footnote{https://sqid.toolforge.org/} and includes a development environment using Jupyter notebooks that provides seamless integration with Pandas \cite{reback2020pandas,mckinney-proc-scipy-2010}. KGTK can process Wikidata-sized KGs, with billions of edges, on a laptop.
Kypher (\texttt{kgtk query}) is one of 55 commands available in KGTK.

\section{Kypher query language and processor}
\label{sec:kypher}

Kypher stands for \emph{KGTK Cypher}. Cypher \cite{francis2018cypher} is a declarative graph query language originally developed at Neo4j.  OpenCypher\footnote{\url{https://opencypher.org/}} is a corresponding open-source development effort for Cypher which forms the basis of the new Graph Query Language (GCL).\footnote{\url{https://www.gqlstandards.org/home}}  We chose Cypher since its ASCII-art pattern language makes it easy even for novices to express complex queries over graph data.

Kypher adopts many aspects of Cypher's query language, but has some important differences. Most notably, KGTK and therefore Kypher do not use the property graph data model assumed by Cypher. Kypher only implements a subset of the Cypher commands (for example, no update commands) and has some minor differences in syntax, for example, to support naming and querying over multiple graphs. Kypher also does not yet support certain features such as path-range patterns, subqueries or unions which are planned as future extensions.

To implement Kypher queries, we translate them into SQL and execute them on SQLite, a lightweight file-based SQL database. Kypher queries are designed to look and feel very similar to other file-based KGTK commands. They take tabular file data as input and produce tabular data as output. There are no servers and accounts to set up, and the user does not need to know that there is in fact a database used underneath to implement the queries. A cache mechanism makes multiple queries over the same KGTK files very efficient. Kypher has been successfully tested on Wikidata-scale graphs with 1.5B edges where queries executing on a standard laptop run in milliseconds to minutes depending on selectivity and result sizes. Additional information about Kypher and its capabilities can be found online. \footnote{\url{https://kgtk.readthedocs.io/en/latest/transform/query/\#overview}}

\section{Use Cases}
\label{sec:usecases}
This section presents five use cases that illustrate different ways to exploit the data in Wikidata. We show how the use cases can be implemented using Kypher queries and executed on a laptop. The equivalent implementations of the queries in SPARQL have been tested against the public Wikidata SPARQL endpoint, and against a large server with a 24 hour time out limit. 
The use cases have been implemented in Python for Kypher (using a Jupyer Notebook) and  SPARQL (using a script) and are available online.\footnote{\url{https://github.com/usc-isi-i2/kgtk-at-2021-wikidata-workshop/}} All input datasets are available in Zenodo under a public  DOI \cite{szekely_pedro_2021_5139550}.
In the paper we illustrate each use case using one Kypher query from the notebook. 
%



\subsection{Retrieval of large amounts of data from Wikidata} 
\label{ssec:names}

John is doing research on the popularity of first names to improve his entity resolution algorithm for people. He sees that Wikidata contains about 9 million people, so he wants to get the distribution of counts of first names from Wikidata. He writes a SPARQL query, but it times out, so he downloads the Wikidata KGTK files on his laptop and writes a Kypher query. The query, shown in Figure~\ref{fig:john}, retrieves all instances of human (Q5), gets their first names using the P735 property (first name) and returns the counts. 

\begin{figure}[hb]
    \begin{verbatim}
!$kypher -i items -i p31 -i labels
--match '
    p31: (person)-[:P31]->(:Q5), # Q5 is person
    items: (person)-[:P735]->(given_name), # P735 is first name
    labels: (given_name)-[:label]->(given_name_label)'
--return 'distinct given_name as node1, count(given_name) as node2, 
given_name_label as `node1;label`, "count_names" as label'
--order-by 'node2 desc'
-o "$OUT"/given-names.tsv

node1      node2   node1;label   label
Q4925477   120416  'John'@en     count_names
Q12344159  74235   'William'@en  count_names
Q4927937   59298   'Robert'@en   count_names
Q16428906  57107   'Thomas'@en   count_names
Q677191    52568   'James'@en    count_names
    \end{verbatim}
    \vspace{-0.2in}
    \caption{Query to count first names in Wikidata (top) and first 5 of 53,253 results (bottom). Runtime: 8.28 minutes.}
    \label{fig:john}
\end{figure}

John chooses standard KGTK names for the headers to generate the data as a KGTK graph so that the output of the query is in the same format as all other KGTK files and can be used as input in future queries. In KGTK, triples are represented using \texttt{node1}, \texttt{label} (i.e., property), and \texttt{node2} headings.
It is also possible to include triples about \texttt{node1} and \texttt{node2} as additional columns by using the semicolon notation. For example, \texttt{node1;label} represents the \texttt{label} property of \texttt{node1} (in KGTK we represent the Wikidata label attribute using the \texttt{label} property).


This type of example computes results over a large number of entities producing a potentially large set of results.
This use case cannot be implemented using the public Wikidata endpoint because the query times out, but the Kypher query runs on a laptop (8 minutes, 16 seconds). The companion Jupyter notebook illustrates how this query can be extended to measure the popularity of first names over time.


\subsection{Analytics on the full Wikidata} 
\label{ssec:class_inst}

Jessica is working with John on the entity resolution algorithm and her job is to use the number of instances of each class in Wikidata as a feature. Jessica just needs to count the number of instances of each class, summing over the instances of all subclasses. She knows that there are over 1 million classes in Wikidata (entities with a \texttt{P279} property), so she knows it will not run on the public SPARQL endpoint. Jessica copies the SQLite database from John so that she does not have to wait the 98 minutes John had to wait to load the needed Wikidata files on her laptop. 

\begin{figure}[tbh]
    \begin{verbatim}
!$kypher -i p31 -i p279star 
--match '
    p31: (entity)-[:P31]->(class), 
    p279star: (class)-[:P279star]->(super_class)' 
--return 'distinct super_class as node1, count(distinct entity) as 
    node2, "entity_count" as label' 
--order-by 'node2 desc, node1' 
-o "$OUT"/class.count.tsv.gz

node1       node2       label
Q35120      88859643    entity_count
Q99527517   74418826    entity_count
Q488383	    73704542    entity_count
\end{verbatim}
    \vspace{-0.2in}
    \caption{Kypher query to count the instances of every class, including instances of the subclasses (top) and the first 3 rows of the result file (bottom). Runtime: 88.97 minutes. }
    \label{fig:jessica}
\end{figure}

The query (Figure~\ref{fig:jessica}) uses two files from the KGTK distribution of Wikidata. The \texttt{p31} file records the class of every instance, and the \texttt{p279star} file records all the super-classes of every class using a new property called \texttt{P279star}. These properties are commonly used so they are provided as separate files for the convenience of users. 
The query retrieves the class from every entity from the \texttt{P31} file, retrieves all the super-classes of every class, and returns the entity count for every super-class.

\begin{figure}[tbh]
    \begin{verbatim}
!$kypher -i p279star -i labels -i "$OUT"/class.count.tsv.gz --as count
--match ' 
    p279star: (class)-[]->(:Q11424), # Q11424 is film
    count: (class)-[:entity_count]->(count), 
    labels: (class)-[:label]->(class_label)' 
--return 'class as node1, class_label as `node1;label`, count as node2' 
--order-by 'cast(count, integer) desc' 
--limit 10

node1      node1;label               node2
Q11424     'film'@en                 314889
Q24862     'short film'@en           33733
Q506240    'television film'@en      17310
    \end{verbatim}
    \vspace{-0.2in}
    \caption{Kypher query to output the counts of all subclasses of file, including indirect subclasses (top) and the top 3 results (bottom). Runtime: 2.4 seconds.}
    \label{fig:jessica2}
\end{figure}


After coming back from lunch, the file is ready. The Kypher query ran in 88.97 minutes on Jessica's laptop, and contains data for 75K classes, as there are many classes that do not have instances. The equivalent SPARQL query timed out on the public SPARQL endpoint and did not complete after 24 hours on our private SPARQL endpoint (see below for the reason why).

Jessica is curious about the data, and wants to know the instance count for all subclasses of film (\texttt{Q11424}). Jessica could modify the query above to include only instances of film with \texttt{p31: (entity)-[:P31]->(:Q11424)}, and the query would return the counts. However, the output file she already computed has all the data she needs, and it is a valid KGTK graph, so Jessica writes a query to pick out the subset that she is interested in (Figure~\ref{fig:jessica2}). She writes a query that uses the output of the previous query (\texttt{class.count.tsv.gz}) as input, uses the \texttt{p279star} file to get all the subclasses of film including indirect subclasses, and fetches the count from the counts file.

The query returns results in 2.4 seconds, illustrating how Kypher makes it easy to chain the results of queries, avoiding recomputation of expensive queries to get results. 
The film version of the query times out on the public SPARQL endpoint and takes 114 seconds on the private SPARQL endpoint as there is no easy way to reuse the results of the previous computation in a new query.
Jessica's Kypher query is efficient because she built a personalized version of Wikidata on her laptop, choosing to add the \texttt{entity\_count} property to her KG to make other queries run quickly. 
Using the file as input to her query was all that Jessica had to do to add the data to her personalized version of Wikidata. Kypher automatically loaded and indexed the data; in addition, Kypher will check whether the files on disk have changed every time it runs a query that uses the file, and will automatically reload and re-index the data as necessary.

This use case is a further example of the previous use case where users want to derive new data from Wikidata using queries that involves a large proportion of the entities in Wikidata. KGTK supports these use cases by making it possible to decompose complex use cases into independent queries, and allowing users to build personalized versions of Wikidata where they extend Wikidata with the results of previous queries. The \texttt{p27star} file that KGTK provides is also an example of this capability, and is the reason why the instances query can run in 80 minutes on a laptop but cannot produce results after 24 hours on a large 256GB server running SPARQL.

\subsection{Extraction of new graphs from Wikidata} 
\label{ssec:author}

Bill is working on a project to find networks of researchers working on specific topics. He wants to use publication data to find relationships among authors. Bill knows that he can get lots of publication data from Pubmed or Microsoft Academic graph, but wants to give Wikidata a try as he heard that Wikidata has close to 40 million publications, and that in Wikidata publications have links to other entities such as main subjects.

\begin{figure}[tbh]
    \begin{verbatim}
!$kypher -i p31 -i p279star -i items -i time -i labels
--match '
    p31: (pub)-[:P31]->(class),
    p279star: (class)-[:P279star]->(:Q591041), # node for scientific publication
    items: (pub)-[:P50]->(author1), # P50 is author
    items: (pub)-[:P50]->(author2)'
--where 'author1 > author2'
--return 'distinct author1 as node1, "Pcoauthor" as label, 
author2 as node2, count(distinct pub) as count_publications'
--order-by 'count_publications desc'
-o "$TEMP"/coauthors.2019.tsv.gz
    \end{verbatim}
    \vspace{-0.2in}
    \caption{Kypher query to build a network of co-authors of publications. Runtime: 66.39 minutes.}
    \label{fig:bill}
\end{figure}

Bill decides that the simplest experiment to try first is to build a network of authors of publications in Wikidata: he wants to create a graph of people in Wikidata who authored papers, to put a link between two people if they coauthored a paper, and to add a qualifier with the count of papers they coauthored. He knows the computation is expensive as there are around 40 million papers in Wikidata, so the network will be large. He doesn't even try to write a SPARQL query because he knows it will time out. Bill downloads the KGTK files and writes the query shown in Figure~\ref{fig:bill}.
The query reuses the \texttt{p31} and \texttt{p279star} files to retrieve all publications that are instances of of any subclass of \texttt{Q591041} (scientific publication). He uses the \texttt{P50} property to retrieve the authors and uses two variables (\texttt{author1} and \texttt{author2}) to retrieve multiple authors if they are present. A \texttt{where} clause ensures that the variables are bound to different authors, and the \texttt{return} clause constructs the output edges and qualifier.
Bill uses the standard \texttt{node1} and \texttt{node2} headings to construct triples, using a new property \texttt{Pcoauthor}. He also invents a new qualifier \texttt{count\_publications} and includes the labels of the authors so that he can read the output.

\begin{figure}[tb]
    \begin{verbatim}
!$kypher -i p31 -i p279star -i items -i labels
--match '
    p31: (pub)-[:P31]->(class),
    p279star: (class)-[:P279star]->(:Q591041), # scientific publication
    items: (pub)-[:P50]->(author1),            # P50 is author
    items: (pub)-[:P50]->(author2),
    items: (pub)-[:P921]->(cancer_type),       # P921 is main subject
    p279star: (cancer_type)-[:P279star]->(:Q12078), # Q12078 is cancer
    labels: (author1)-[:label]->(author1_label),
    labels: (author2)-[:label]->(author2_label)'
--where 'author1 > author2'
--return '
    distinct author1 as node1, "Pcoauthor" as label, author2 as node1,
    count(distinct pub) as count_publications, 
    author1_label as `node1;label`, author2_label as `node2;label`'
--order-by 'count_publications desc'
-o "$TEMP"/coauthors.cancer.tsv.gz

node1      label      node2    count  node1;label     node2;label
Q60320900  Pcoauthor  Q60394812  396  'Jorge ...'@en  'Hagop Kan ...'@en
Q60394812  Pcoauthor  Q66370727  236  'Hagop ...'@en  'Susan O'Brien'@en
Q40614280  Pcoauthor  Q60394812  186  'Farha ...'@en  'Hagop Kan ...'@en
    \end{verbatim}
    \vspace{-0.2in}
    \caption{Kypher query to build a network of co-authors of publications about cancer (top) and the top 3 results (bottom). Runtime: 2.62 minutes}
    \label{fig:bill2}
\end{figure}

Bill continues his investigation. He is interested in cancer research, so he wants to build the same network but using only the papers about cancer. 
He extends the query using the \texttt{P921} property to retrieve the main subjects of a paper, and again uses the \texttt{p279star} file to select subjects that are below cancer (\texttt{Q12078}).
He expects the query to be much faster because now it has strong restrictions, so he gives it a try. The query, shown in Figure~\ref{fig:bill2}, takes 2.62 minutes and produces a network with close to half a million edges.


\subsection{Queries combining multiple resources}
\label{ssec:ulan}

Abigail is working on a cultural heritage project, collaborating with the Getty Research Institute who gave her a file with 27 thousand identifiers of artists that she is interested in; the file has one identifier per line. The Getty uses ULAN identifiers\footnote{\url{https://www.getty.edu/research/tools/vocabularies/ulan/}}, and Abigail has a database indexed using VIAF identifiers.\footnote{\url{http://viaf.org}} Abigail needs to map the ULAN identifiers to VIAF identifiers so that she can use her database. She puts one of the ULAN identifiers in the Wikidata search box and discovers that Wikidata has both ULAN and VIAF identifiers for many artists, so she needs to write a query that retrieves artists using the ULAN identifiers and returns the VIAF identifier when it is available in Wikidata.

Abigail considers using SPARQL as it is easy to write a query to retrieve the VIAF identifier given a ULAN identifier. This solution would require sending 27,000 queries to Wikidata (or 27 queries binding 1000 identifiers), and would involve writing a script.
She writes a SPARQL query that binds all 27,000 identifiers, but the query is too large and it is rejected in the public SPARQL endpoint.

\begin{figure}[tb]
    \begin{verbatim}
!$kypher -i items -i external_ids -i labels -i "$OUT"/ulan.tsv 
--match '
    ulan: (ulan_id)-[]->(), 
    # P214 is VIAF ID, P245 is Union List of Artist Names ID
    external_ids: (viaf_id)<-[:P214]-(artist)-[:P245]->(ulan_id), 
    labels: (artist)-[]->(artist_label)' 
--return '
    artist as node1, viaf_id as node1;P214, ulan_id as node1;P245, 
    artist_label as node1;label' 
-o "$OUT"/ulan-to-viaf.tsv

node1       node1;P214    node1;P245  node1;label
Q1000596    "20822441"   "500072302"  'Noémi Ferenczy'@en
Q1001063    "96418002"   "500099612"  'Olga Fialka'@en
Q100156272  "309815799"  "500335625"  'Gloria López Córdova'@en
    \end{verbatim}
    \vspace{-0.2in}
    \caption{Kypher query to retrieve VIAF identifiers for a file with 27,000 ULAN identifiers. Runtime: 11.8 seconds.}
    \label{fig:abigail}
\end{figure}

Abigail solves the problem using Kypher. She first renames the heading of the \texttt{ulan.tsv} file to \texttt{node1}, so the file is now a valid KGTK graph file because in KGTK any value can be used as \texttt{node1} (subject). Her ULAN KGTK graph contains 27,000 nodes and no edges for any node. 
She writes the Kypher query shown in Figure~\ref{fig:abigail}. The query uses the \texttt{ulan.tsv} file as input, thereby personalizing the Wikidata graph to include the ULAN identifier nodes.
The query binds the \texttt{ulan\_id} variable to the nodes in the \texttt{ulan.tsv} graph.
The next clause  uses the \texttt{external\_ids} graph from the KGTK distribution of Wikidata to map the ULAN ids to VIAF ids.
She returns the data by using standard KGTK headers so that she can use the resulting file in other queries.
The query runs in 11.8 seconds and retrieves 8,116 VIAF ids.

Wikidata has become a hub for identifiers as it contains a large number of identifiers for entities (over 160 million identifiers).
Abigail's use case is an example of a common use case to exploit the Wikidata identifiers: a researcher has an external source that contains identifiers present in Wikidata and wants to retrieve the entities or map one type of identifier to another. 
Kypher queries address this use case as external resources can be easily converted to KGTK graphs and used in queries.

\subsection{Combination of Wikidata and DBpedia}
\label{ssec:dbpedia}

After mining her VIAF database, Abigail realizes that she needs more data and wants to exploit the Wikipedia infoboxes. 
Abigail considers using SPARQL federated queries~\cite{quilitz2008querying} to combine Wikidata and DBpedia, but she faces the same problem as before in that she has 27,000 identifiers and may need to issue a large number of queries.
Abigail downloads the DBpedia infobox data in RDF format from the DBpedia Databus\footnote{\url{https://databus.dbpedia.org/dbpedia/}} and uses KGTK commands to convert the data to KGTK format.
The resulting KGTK file contains close to 100 million edges but the data is noisy as illustrated in the following excerpt.
Abigail expects the \texttt{node2} column for these properties to contain Wikidata q-nodes, but sees that often, literals are present.
{\footnotesize
\begin{verbatim}
        node1    label                 node2
        Q466241	 property:almaMater	   Q2746779
        Q466241	 property:occupation   'Fashion designer'@en
        Q466241	 property:spouse       'Patrick Robyn'@en
\end{verbatim}
}

\begin{figure}[tbh]
    \begin{verbatim}
!$kypher -i infobox -i p31 -i labels
--match '
    infobox: (artist)-[:`property:spouse`]->(spouse),
    p31: (spouse)-[]->(:Q5)'
--opt 'labels: (spouse)-[:label]->(spouse_label)'
--return 'artist as node1, "P26" as label, spouse as node2, 
          spouse_label as `node2;label`'
-o "$OUT"/spouses.dbpedia.qnodes.tsv

node1     label  node2     node2;label
Q268177   P26    Q1000505  'Bud Lee'@en
Q673856   P26    Q1000682  'Fernando Carrillo'@en
Q1325720  P26    Q1000874  'Thomas Montacute, 4th Earl of ...'@en
    \end{verbatim}
    \vspace{-0.2in}
    \caption{Kypher query to retrieve spouse statements from DBpedia and verify that the spouses are instances of human in Wikidata. Runtime: 3.4 minutes.}
    \label{fig:abigail-db}
\end{figure}

Abigail is interested in the spouse data for her artists and writes the query shown in Figure~\ref{fig:abigail-db}. The query uses as input an external file (\texttt{infobox}) and a Wikidata file (\texttt{p31}),  retrieves the spouse from the DBpedia file and verifies that the value of spouse is and instance of human (\texttt{Q5}).

Kypher loads the 100 million edge DBpedia file in 10.6 minutes and runs the query in $3.4$ minutes, a fast time considering that there are $322,599$ spouse edges in the DBpedia graph, and all must be checked to be instances of human (\texttt{Q5}).
The query identifies $7,325$ high quality spouse statements in DBpedia infoboxes that are not present in Wikidata. 

This use case illustrates building a personalized Wikidata extension that augments Wikidata with noisy data and then uses Kypher to extract clean data. 


\section{Experiments}
\label{sec:experiments}

\begin{table}[b]
    \centering
    \label{tab:experiments}
    \caption{Comparison of execution times (minutes) of queries in use cases. (*) submitting over $5,000$ ULAN identifiers produces an error due to the length of the query.}
    \label{tab:time}
    \begin{tabular}{| l | c | c | c | c |}
        \hline
         \bf{Query} & \bf Kypher & \bf Kypher & \bf SPARQL & \bf SPARQL \\ 
                    & \bf 16GB & \bf 32GB & \bf local (256GB)& \bf public \\ \hline
        First names & 24.37 & 8.28 & 31.05 & time out\\
        Class instances & 104.97 & 88.97 & $>$24 hours & time out  \\
        Film instances & 0.03 & 0.04 & 1.91 & time out  \\
        Author network & 61.55 & 66.39 & $>$24 hours &  time out\\
        Cancer network & 3.18 & 2.62 & 40.19& time out \\
        ULAN identifiers & 0.56 & 0.20 & 1.08 & * \\
        DBpedia spouses & 3.92 & 3.43 & n/a & n/a \\ 
        \hline
    \end{tabular}
    
\end{table}

We performed experiments to compare the execution times of the Kypher and SPARQL implementations of the queries for the use cases presented in this paper. We used two configurations for Kypher, MacBook Pro laptops with 16GB memory/256GB SSD disk and 32 GB memory/1TB SSD. We used two configurations for SPARQL, the public Wikidata SPARQL endpoint 
and a local clone of Wikidata (June 2019) running on a server with 24 Intel Xeon cores and 256 GB of memory and SSD. 
For the Kypher queries we used the Wikidata February 15, 2021 distribution converted to KGTK format. For the Wikidata clone we used the RDF dump from the June 15, 2019 distribution (we did not load the February 15, 2021 distribution as it takes several days to load, and for the purpose of our experiments the earlier dump is adequate as it is smaller).

The Jupyter notebook for the Kypher queries was run twice. The first run of the notebook with an empty SQLite database took $349$ minutes in the 32GB laptop, and the second run of the notebook, after the data was loaded and indices were built, took $164$ minutes. The difference, $185$ minutes includes $98$ minutes to load the Wikidata data, $10$ minutes to load the DBpedia infobox data, and the rest, $77$ minutes is time that Kypher used to build database indices. 



Table~1 shows the runtimes of the queries presented in the use cases. The times are from the second run of the notebook after the data was loaded in the Kypher SQLite database and indices had been created.

\section{Discussion and Conclusions}
\label{sec:discussion}
The main objective of KGTK and Kypher is to democratize the exploitation of Wikidata so that anyone with modest computing resources can take advantage of the vast amounts of knowledge present in Wikidata. Our tools focus on use cases that use large portions of Wikidata to distill new knowledge. The experiments show how expensive queries (e.g., class instances use case, Section~\ref{ssec:class_inst}) that cannot run in one day on a powerful server, complete in about one hour on a laptop. Analytic queries (first names use case, Section~\ref{ssec:names}) become possible on a laptop in a few minutes, and distillation of knowledge for analysis (author and cancer network use case, Section~\ref{ssec:author}) become practical. Kypher enables users to easily combine Wikidata with external sources to extract relevant Wikidata knowledge (ULAN use case, Section~\ref{ssec:ulan}), or to enhance Wikidata with knowledge from external sources (DBpedia use case, Section~\ref{ssec:dbpedia}). 

Kypher is not meant to address use cases that require the most up-to-date data in Wikidata. KGTK uses the Wikidata JSON dumps published every few days, and the KGTK workflow to process the JSON dump takes one day.   

The comparison with the Wikidata SPARQL endpoints is preliminary as we have not controlled for caching in the triple store and in the operating system, or performed systematic variations of the complexity of the queries. A more detailed and controlled comparison is reserved for a future paper.  Here we speculate on the reasons why Kypher seems to perform significantly better than the Wikidata SPARQL endpoints on the presented use cases:

\begin{enumerate}
    \item \textit{Compact data model:} the KGTK data model allows us to translate 1.2B Wikidata statements very directly into 1.2B edges, while the RDF translation requires reification and generates \textit{O}(10B) triples.  KGTK also does not require the use of namespaces which makes data values more compact.
    \item \textit{Smaller database size:} more compact data translates directly into smaller database sizes, for example, 142GB for the Kypher graph cache vs. 718GB for the local Wikidata endpoint.  This gives generally better locality for table and index lookups and better caching of data pages.
    \item \textit{Specialized tables:} representing specialized data slices such as \texttt{P279star} in their own graph tables makes their reuse very efficient and their indexes more focused, compact, and cache-friendly.
    \item \textit{Read-only processing:} Kypher does not need to support fine-grained updates of tables and indexes, which need to be supported by the public Wikidata endpoint. This requires additional machinery that slows down performance.
    \item \textit{Use case selection:} triple stores and databases are optimized to support a large number of use cases. Our set of use cases samples a small slice of that space, and  performance might be very different for other types of queries. 
\end{enumerate}

The contribution of this paper is to show that KGTK and Kypher are effective tools for complex analytic use cases.
The paper demonstrates that Kypher supports a variety of use cases cases that are impractical with existing tooling. Kypher allows researchers and developers to investigate use cases on their own laptop, exploring extensions of Wikidata that would not be possible on shared resources, with minimal setup, using a simple query language.

{\bf Acknowledgements:} This material is based on research sponsored by Air Force Research Laboratory under agreement number FA8750-20-2-10002. The U.S. Government is authorized to reproduce and distribute reprints for Governmental purposes notwithstanding any copyright notation thereon. The views and conclusions contained herein are those of the authors and should not be interpreted as necessarily representing the official policies or endorsements, either expressed or implied, of Air Force Research Laboratory or the U.S. Government.

 \bibliographystyle{splncs04}
 \bibliography{refs}

\end{document}